\documentclass[conference]{IEEEtran}

\usepackage{epsf,psfig}

\usepackage{algorithmic,algorithm} 
\usepackage{amsmath}

\newcommand{\commentaire}[1]{ } 
\newcommand{\definition}[1]{\textbf{Definition:} #1\\} 

\title{Scuba Search : when selection meets innovation}

\author{\authorblockN{S\'ebastien Verel}
\authorblockA{Universit\'{e} de Nice-Sophia Antipolis, \\
06640, Laboratoire I3S, France, \\
Email: verel@i3s.unice.fr}
\and
\authorblockN{Philippe Collard}
\authorblockA{Universit\'{e} de Nice-Sophia Antipolis, \\
06640, Laboratoire I3S, France, \\
Email: pc@i3s.unice.fr}
\and
\authorblockN{Manuel Clergue}
\authorblockA{Universit\'{e} de Nice-Sophia Antipolis, \\
06640, Laboratoire I3S, France, \\
Email: clerguem@i3s.unice.fr}
}

\begin{document}

\maketitle

\begin{abstract}
We proposed a new search heuristic using the \textit{scuba diving} metaphor. This approach is based on the concept of evolvability and tends to exploit neutrality in fitness landscape. Despite the fact that natural evolution does not directly select for evolvability, the basic idea behind the \textit{scuba search} heuristic is to explicitly push the evolvability to increase. The search process switches between two phases: \textit{Conquest-of-the-Waters} and \textit{Invasion-of-the-Land}. A comparative study of the new algorithm and standard local search heuristics on the NKq-landscapes has shown advantage and limit of the scuba search. To enlighten qualitative differences between neutral search processes, the space is changed into a connected graph to visualize the pathways that the search is likely to follow.

\end{abstract}

\section{Introduction}
In this paper we propose a novel heuristic called \textit{Scuba Search} that allows to exploit the neutrality that is existing in many real-world fitness landscapes.
\par
This section presents the interplay between neutrality in fitness landscapes 
and metaheuristics. Section II describes the {\it Scuba Search} heuristic 
in detail. In order to illustrate the efficiency of this heuristic, we 
use the $NKq$-landscape as a model of neutral fitness landscape. This one is developped 
in section III. The experiment results are given in section IV where 
comparisons are made with standard heuristics. Section V analyzes 
the neutral search process of scuba search. We point out in section VI 
the shortcoming of the approach and propose a {\it generic scuba 
search} heuristic. Finally, we summarize our contribution and present 
plans for future work.

\subsection{Neutrality}

The metaphor of 'adaptative landscape' introduced by S.~Wright~\cite{wright:rmicse} has dominated the view of adaptive evolution: an 
uphill walk of a population on a mountainous fitness landscape in which it can get stuck on suboptimal peaks.
Results given by molecular evolution has changed this picture: Kimura~\cite{KIM:83} establishes that the overwhelming majority of mutations are either effectively neutral or lethal and in the latter case purged by negative selection. This theory is called the theory of molecular evolution. 
This theory can help us to revisit the metaphor of adaptive landscape and define a neutral landscape. In neutral landscape population can walk on mountain but also on plateaus (neutral networks). The dynamics of population evolution is then a metastable evolution as proposed by Gould and Eldredge \cite{GOU:77} characterized by long periods of fitness stasis (population stated on a 'neutral network') punctuated by shorter periods of innovation with rapid fitness increase.
\commentaire{
Under this hypothesis, dynamics of populations evolving on such 
neutral landscapes are different from those on adaptive landscapes: they 
are characterized by long periods of fitness stasis (population stated on 
a 'neutral network') punctuated by shorter periods of innovation with rapid fitness 
increase.}
\par
In the field of evolutionary computation, neutrality plays an important role. Under the assumption that the neutral networks are nearly isotropic\footnote{same neutral degree and same probability to jump}, Barnett~\cite{barnett01netcrawling} proposes an heuristic adapted to neutral landscape: the {\it Netcrawler process}. It is a random neutral walk with a mutation mode adapted to local neutrality. The per-sequence mutation rate is optimized to jump to one neutral network to another one. 
When the isotropic assumption is not verified, for a population based algorithm, Nimwegen {\it et al.}~\cite{NIM:99} show the population's limit distribution on the neutral network is solely determined by the network topology. The population seeks out the most connected areas of the neutral network.
\par
In real-world problems as in design of digital circuits 
\cite{vassilev00advantages}\cite{thompson00evolution}\cite{layzell98evolvable}, in evolutionary robotics \cite{jakobi95noise} or in genetic programming \cite{YU:01}, neutrality is implicitly embedded in the genotype to phenotype mapping. Another possibility in evolutionary 
optimization is to introduce artificial redundancy into the encoding 
\cite{KNO:02}\cite{collard99synthetic}. This may improve the evolvability of genotype or create 
neutral paths to escape from suboptimal peaks.

\subsection{Evolvability}

\textit{Evolvability} is defined by Altenberg~\cite{WA-AL} as "the ability of random variations to sometimes produce improvement". This concept refers to the efficiency of evolutionary search; it is based upon the work by Altenberg~\cite{ALT:94}: "the ability of an operator/representation scheme to produce offspring that are fitter than their parents". Smith et al.~\cite{SMI:01} focus on the ideas of \textit{evolvability} and \textit{neutrality}; they plot the average fitness of offspring over fitness of parents (considering 1-bit mutation as an operator). As enlighten by Turney~\cite{turney99increasing} the concept of evolvability is difficult to define. As he puts it: "if $s$ and $s{'}$ are equally fit, $s$ is more \textit{evolvable} than $s{'}$ if the fittest offspring of $s$ is more likely to be fitter than the fittest offspring of $s{'}$". Following this idea we define evolvability as a function (see section \ref{def_evol}). 




\section{Scuba Search}
\subsection{The conquest of the waters}
Keeping the landscape as a model, fill each area between two peaks (local
optima) with water allow lakes to emerge. Thus, the landscape is bathed in
an uneven sea; areas under water represent non-viable solutions. So now
there are paths from one peak to the other one for a swimmer. The key, of
course, remains to locate an attractor which represents the system's maximum
fitness. In this framework, the problem is how to cross a lake without global
information. We use the scuba diving metaphor as a guide to present the 
principles of the so-called \textit{scuba search} ($SS$). This heuristic is a
way to deal with the problem of crossing between peaks and so avoid to be
trapped in the vicinity of local optima. The problem is what drives the swimmer
from one edge to the opposite edge of the lake? The classic view is the one
of a swimmer drifting at the surface of a lake. The new metaphor is a scuba
diver seeing the \textit{world above the water surface}. We propose a new heuristic to cross a neutral net getting information above-the-surface (ie. from fitter points of the neighborhood).

\subsection{Scuba Search Algorithm}

Despite the fact that natural evolution does not directly select for evolvability, there is a dynamic pushing evolvability to increase~\cite{turney99increasing}. As Dawkins~\cite{Dawkins-1996} states, "This is not ordinary Darwinian selection but it is a kind of high-level analogy of Darwinian selection". The basic idea behind the $SS$ heuristic is to explicitly push evolvability to increase. Before presenting this search algorithm, we need to introduce a new type of local optima, the {\it local-neutral optima}. Indeed with $SS$ heuristic, local-neutral optima will allow transition from neutral to adaptive evolution. So evolvability will be locally optimized.\\

Given a search space ${\cal S}$ and a fitness function $f$ defined on ${\cal S}$, some more precise definitions follow.

\definition{A {\it neighborhood structure} is a function ${\cal V} : {\cal S} \rightarrow 2^{\cal S}$ that assigns to every $s \in {\cal S}$ a set of neighbors ${\cal V}(s)$ such that $s \in {\cal V}(s)$}

\label{def_evol}
\definition{The {\it evolvability} of a solution $s$ is the function $evol$ that assigns to every $s \in {\cal S}$ the maximum fitness from the neighborhood ${\cal V}(s)$: $\forall s \in {\cal S}$, $evol(s) = max \lbrace f(s^{'}) ~|~ s^{'} \in  {\cal V}(s) \rbrace$}
\commentaire{Definition : A {\it local maximum} is a solution $s_{opt}$ such that $ \forall s^{'} \in {\cal V}(s_{opt})$, $ f(s^{'}) \leq f(s_{opt})$. We call $s_{opt}$ a {\it strict local maximum} if $ \forall s^{'} \in {\cal V}(s_{opt})$, $ f(s^{'}) < f(s_{opt})$\\
}

\definition{For every fitness function $g$, neighborhood structure ${\cal W}$ and genotype $s$, the predicate $isLocal$ is defined as:\\
$isLocal(s, g, {\cal W}) = (\forall s^{'} \in {\cal W}(s), g(s^{'}) \leq g(s) )$
}

\definition{For every $s \in {\cal S}$, the {\it neutral set} of $s$ is the set ${\cal N} (s) = \lbrace s^{'} \in {\cal S}~|~ f(s^{'}) = f(s) \rbrace$, and the {\it neutral neighborhood} of $s$ is the set ${\cal V}n(s) = {\cal V}(s) \cap {\cal N}(s)$}

\definition{For every $s \in {\cal S}$, the {\it neutral degree} of $s$, noted $Degn(s)$, is the number of neutral neighbors of $s$, $Degn(s) = \#{\cal V}n(s) - 1$}

\definition{A solution $s$ is a \textit{local maximum} iff $isLocal(s, f, {\cal V})$}

\definition{A solution $s$ is a \textit{local-neutral maximum} iff $isLocal(s, evol, {\cal V}n)$}

There are two overlapping dynamics during the Scuba Search process. The first one is identified as a neutral path. At each step the scuba diving remains under the water surface driven by the hands-down fitnesses; that is fitter fitness values reachable from one neutral neighbor. At that time the \textit{flatCount} counter is incremented. When the diving reaches a local-neutral optimum, that is if all the fitnesses reachable from one neutral neighbor are selectively neutral or disadvantageous, the neutral path stops and the diving starts up the \textit{Invasion-of-the-Land}. Then the \textit{gateCount} counter increases. This process goes along, switching between \textit{Conquest-of-the-Waters} and \textit{Invasion-of-the-Land}, until a local optimum is reached.

\begin{algorithm}
\caption{Scuba Search}
\label{algoScuba}

\begin{algorithmic}
\STATE flatCount $\leftarrow$ 0, gateCount $\leftarrow$ 0
\STATE Choose initial solution $s \in \cal S$
\REPEAT

	\WHILE{not $isLocal(s, evol, {\cal V}n)$}
		\STATE $M = max \lbrace evol(s^{'})~|~s^{'} \in {\cal V}n(s)-\{s\} \rbrace$
		\IF{$evol(s) < M$}
			\STATE choose $s^{'} \in {\cal V}n(s)$ such that $evol(s^{'}) = M$
			\STATE $s \leftarrow s^{'}$, flatCount $\leftarrow$ flatCount +1
		\ENDIF
	\ENDWHILE
	\STATE choose $s^{'} \in {\cal V}(s) - {\cal V}n(s)$ such that $f(s^{'}) = evol(s)$
	\STATE $s \leftarrow s^{'}$, gateCount $\leftarrow$ gateCount +1
\UNTIL{$isLocal(s, f, {\cal V})$}
\end{algorithmic}
\end{algorithm}

\section{Model of Neutral Landscape}

\commentaire{
In order to study the Scuba Search heuristic we use $NKq$-landscape~\cite{newman98effect}. $NKq$ is a benchmark with an tunable degree of neutrality inspired by $NK$ fitness landscape. Parameter $K$ can tune the ruggedness and parameter $q$ the neutrality of the landscape.}

In order to study the Scuba Search heuristic we have to use landscapes with a tunable degree of neutrality.
The $NKq$ fitness landscapes family proposed by Newman {\it et al}.~\cite{newman98effect} has properties of systems undergoing neutral selection such as RNA sequence-structure maps. 
It is a generalization of the $NK$-landscapes proposed by Kauffman~\cite{KAU:93} where parameter $K$ can tune the ruggedness and parameter $q$ the degree of neutrality of the landscape.


\subsection{Definition}

The fitness function of a $NKq$-landscape is a function $f: \lbrace 0, 1 \rbrace^{N} \rightarrow [0,1]$ defined on binary strings with $N$ loci.
Each locus $i$ represents a gene with two possible alleles, $0$ or $1$. 
An 'atom' with fixed epistasis level is represented by a fitness components $f_i: \lbrace 0, 1
\rbrace^{K+1} \rightarrow [0,q-1]$ associated to each locus $i$. It depends
on the allele at locus $i$ and also on the alleles at $K$ other epistatic loci
($K$ must fall between $0$ and $N - 1$). The fitness $f(x)$ of $x \in \lbrace 0, 1 \rbrace^{N}$
is the average of the values of the $N$ fitness components $f_i$:\label{defNK}

$$ f(x) = \frac{1}{N (q-1)} \sum_{i=1}^{N} f_i(x_i; x_{i_1}, \ldots, x_{i_K})
$$ where $\lbrace i_1, \ldots, i_{K} \rbrace \subset \lbrace 1, \ldots, i -
1, i + 1, \ldots, N \rbrace$. Many ways have been proposed to choose the $K$ other
loci from $N$ loci in the genotype. Two possibilities are mainly used: adjacent and
random neighborhoods. With an adjacent neighborhood, the $K$ genes
nearest to the locus $i$ are chosen (the genotype is taken to have periodic boundaries).
With a random neighborhood, the $K$ genes are chosen randomly on the genotype.
Each fitness component $f_i$ is specified by extension, ie an integer number $y_{i,(x_i; x_{i_1}, \ldots, x_{i_K})}$ from $[0, q - 1]$ 
is associated with each element $(x_i; x_{i_1}, \ldots, x_{i_K})$ from $\lbrace 0, 1 \rbrace^{K+1}$.
Those numbers are uniformly distributed in the interval $[0, q - 1]$.


\subsection{Properties}

The parameters of $NKq$-landscape tune ruggedness and neutrality of the landscape~\cite{newman98effect}\cite{nic-comparison}. The number of local optima is link to parameter $K$. The largest number is obtained when $K$ takes its maximum value $N-1$. The neutral degree (see tab. \ref{tab_Degn}) decreases when $q$ increases and when $K$ increases. The maximal degree of neutrality appears when $q$ takes the value $2$.

\begin{table}
\caption{Average neutral degree on $NKq$-landscape with $N=64$ performs on $50000$ genotypes}
\begin{center}
\begin{tabular}{|c||c|c|c|c|c|c|}
\hline
 & \multicolumn{6}{|c|}{K} \\ 
\cline{2-7}
q & 0 & 2 & 4 & 8 & 12 & 16 \\
\hline \cline{2-7}
2 & 35.00 & 21.33 & 16.56 & 12.39 & 10.09 & 8.86 \\
3 & 21.00 & 13.29 & 10.43 & 7.65 & 6.21 & 5.43 \\
4 & 12.00 & 6.71 & 4.30 & 2.45 & 1.66 & 1.24 \\
100 & 1.00 & 0.32 & 0.08 & 0.00 & 0.00 & 0.00 \\
\hline
\end{tabular}	
\end{center}
\label{tab_Degn}
\end{table}

\section{Experiment Results}

\subsection{Algorithm of Comparison}
\label{sec_algo}
Three algorithms of comparison are used: two kinds of {\it Hill 
Climbing} and one heuristic adapted to neutral landscape, the {\it Netcrawler Process}.

\subsubsection{Hill Climbing}
The simplest type of local search is known as \textit{Hill Climbing} ($HC$) when trying to maximize a solution. $HC$ is very good at exploiting the neighborhood; it always takes what looks best at that time. But this approach has some problems. The solution found depends from the initial solution. Most of the time, the found solution is only a local optima.
We start off with a probably suboptimal solution. Then we look in the neighborhood of that solution to see if there is something better. If so, we adopt this improved solution as our current best choice and repeat. If not, we stop assuming the current solution is good enough (local optimum).
\begin{algorithm}
\caption{Hill Climbing}
\label{algoHC}
\begin{algorithmic}
\STATE step $\leftarrow$ 0
\STATE Choose initial solution $s \in \cal S$
\REPEAT
	\STATE choose $s^{'} \in {\cal V}(s)$ such that $f(s^{'}) = evol(s)$
	\STATE $s \leftarrow s^{'}$, step $\leftarrow$ step + 1
\UNTIL{$isLocal(s, f, {\cal V})$}
\end{algorithmic}
\end{algorithm}

\subsubsection{Netcrawler Process}
We also compare $SS$ to \textit{Netcrawler Process} ($NC$) proposed by Barnett~\cite{barnett01netcrawling}. This is a local search adapted to fitness landscapes featuring neutral networks\footnote{More exactly to $\epsilon$-correlated fitness landscapes}. Netcrawler uses a mutation per-sequence mutation rate which is calculated from the neutral degree of neutral networks~\cite{barnett01netcrawling}. In the case of $NKq$-landscapes, experimentations and estimations of neutral degree given by \cite{nic-comparison} yields as mutation flips only one bit per genotype. The algorithm \ref{algoNetcrawler} displays the Netcrawler process.

\begin{algorithm}
\caption{Netcrawler Process}
\label{algoNetcrawler}
\begin{algorithmic}
\REQUIRE stepMax $ > 0$
\STATE step $\leftarrow 0$
\STATE Choose initial solution $s \in \cal S$
\REPEAT
\STATE choose $s^{'} \in {\cal V}(s)$ randomly 
\IF{$f(s) \leq f(s^{'})$}
	\STATE $s \leftarrow s^{'}$
\ENDIF
\STATE step $\leftarrow$ step $+ 1$
\UNTIL{stepMax $ \leq $ step}
\end{algorithmic}
\end{algorithm}

\subsubsection{Hill Climbing Two Steps}

Hill Climber can be extended in many ways. {\it Hill Climber two Step} ($HC2$) exploits a larger neighborhood of stage 2. The algorithm is nearly the same as $HC$. $HC2$ looks in the extended neighborhood of stage two of the current solution to see if there is something better. If so, $HC2$ adopts the solution in the neighborhood of stage one which can reach a best solution in the extended neighborhood. If not, $HC2$ stop assuming the current solution is good enough.
So, $HC2$ can avoid more local optimum than $HC$. 
Before presenting the algorithm \ref{algoHCDeuxPas} we must introduce the following definitions:

\definition{The {\it extended neighborhood structure\footnote{Let's note that ${\cal V}(s) \subset {\cal V}^2(s)$}} from ${\cal V}$ is the function ${\cal V}^2(s) = \cup_{s_1 \in {\cal V}(s)} {\cal V}(s_1)$}

\definition{$evol^2$ is the function that assigns to every $s \in {\cal S}$ the maximum fitness from the extended neighborhood ${\cal V}^2(s)$. $\forall s \in {\cal S}$, $evol^2(s) = max \lbrace f(s^{'}) | s^{'} \in {\cal V}^2(s) \rbrace$}

\begin{algorithm}
\caption{Hill Climbing (Two Steps)}
\label{algoHCDeuxPas}
\begin{algorithmic}
\STATE step $\leftarrow$ 0
\STATE Choose initial solution $s \in \cal S$
\REPEAT
\IF{$evol(s)=evol^2(s)$}
	\STATE choose $s^{'} \in {\cal V}(s)$ such that $f(s^{'}) = evol^2(s)$
\ELSE
	\STATE choose $s^{'} \in {\cal V}(s)$ such that $evol(s^{'}) = evol^2(s)$
\ENDIF
\STATE $s \leftarrow s^{'}$, step $\leftarrow$ step + 1
\UNTIL{$isLocal(s, f, {\cal V}^2)$}
\end{algorithmic}
\end{algorithm}


\subsection{Parameters setting}

All the four heuristics are applied to a same instance of $NKq$ fitness landscape\footnote{With random neighborhood}. The search space ${\cal S}$ is $\lbrace 0, 1 \rbrace^{N}$, that is bit strings of length $N$. In this paper all experiments are led with $N=64$. The selected neighborhood is the classical one-bit mutation neighborhood: ${\cal V}(s) = \lbrace s^{'}~|~ Hamming(s^{'}, s) \leq 1 \rbrace$. For each triplet of parameters $N$, $K$ and $q$, $10^3$ runs were performed. For netcrawler process, stepMax is set to $300$\footnote{In our experiments netcrawler stops moving before this limit}.

\subsection{Average performances}
\begin{figure*}[!tb] 
\begin{center}
\begin{tabular}{cc} 
\psfig{figure=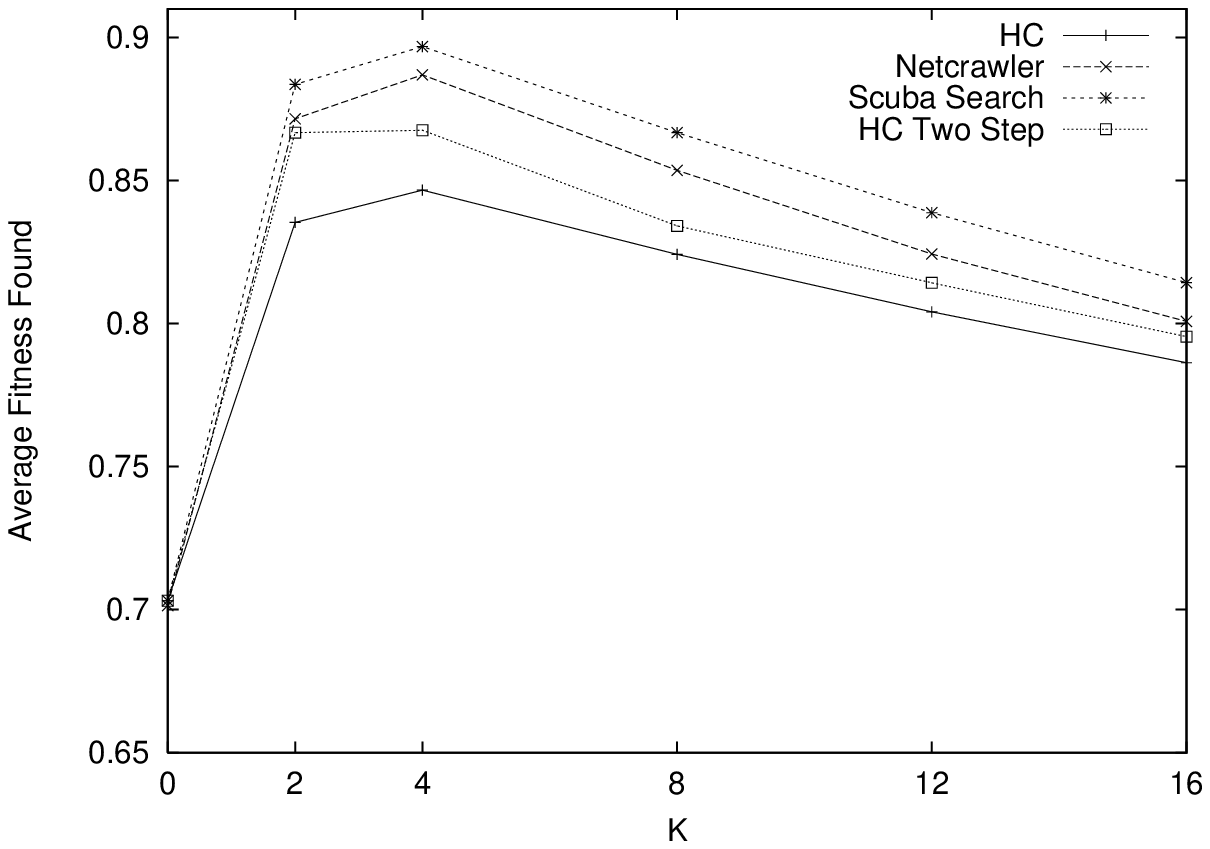,width=230pt,height=180pt} 
&
\psfig{figure=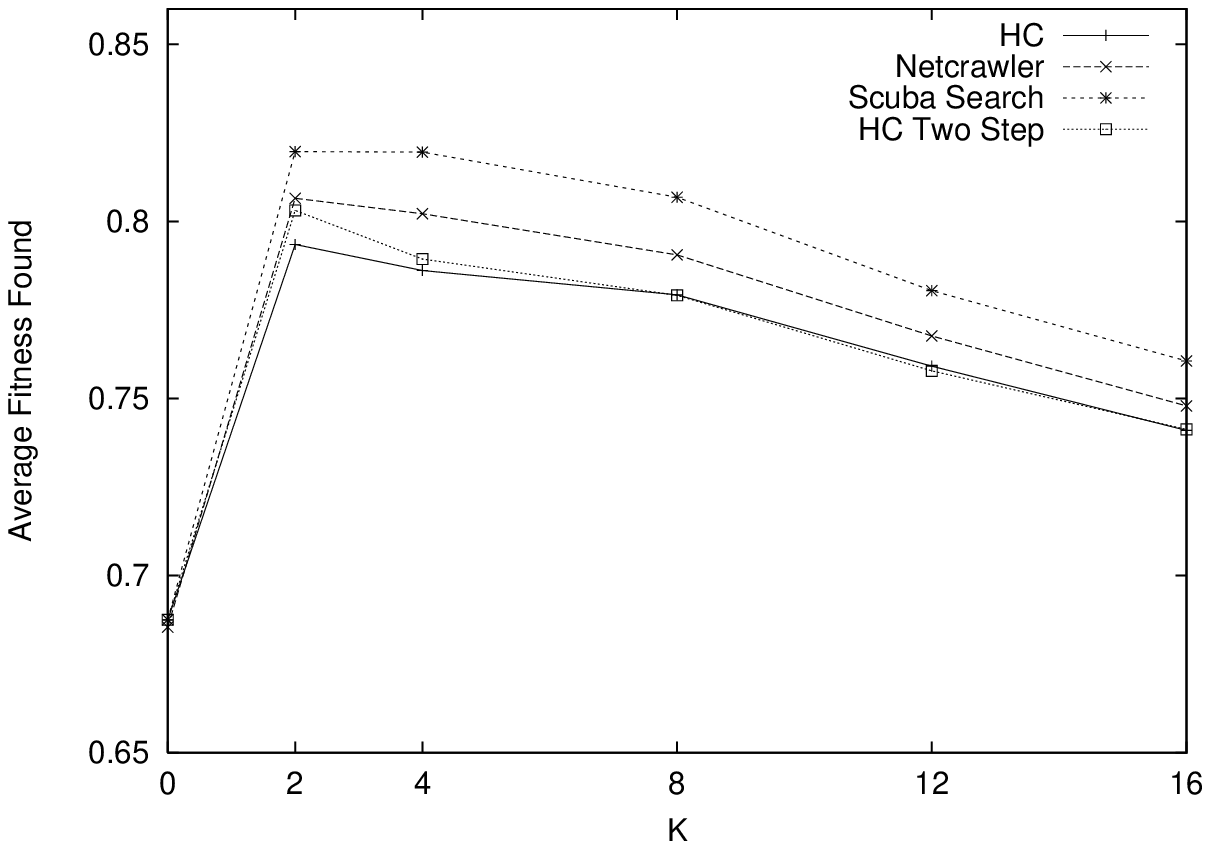,width=230pt,height=180pt}
\\
(a) &
(b) \\
\psfig{figure=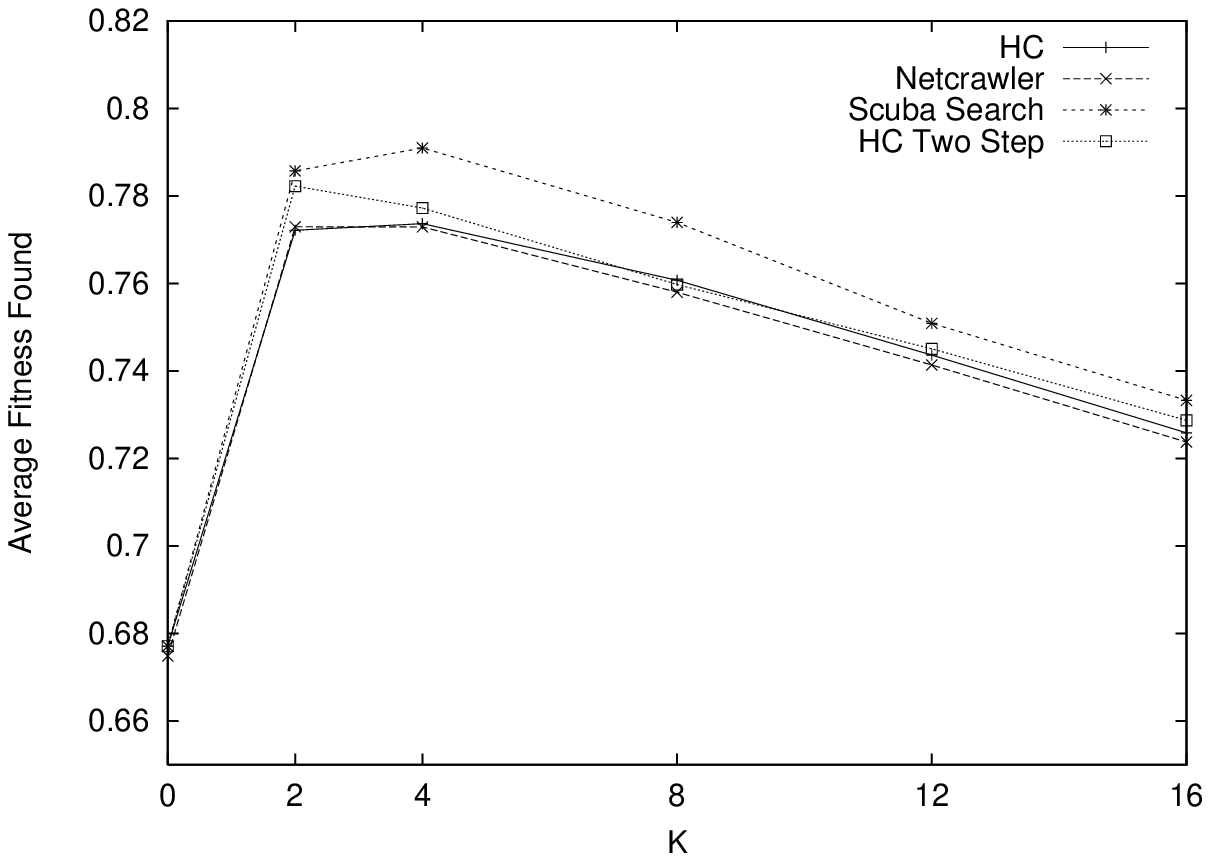,width=230pt,height=180pt}
&
\psfig{figure=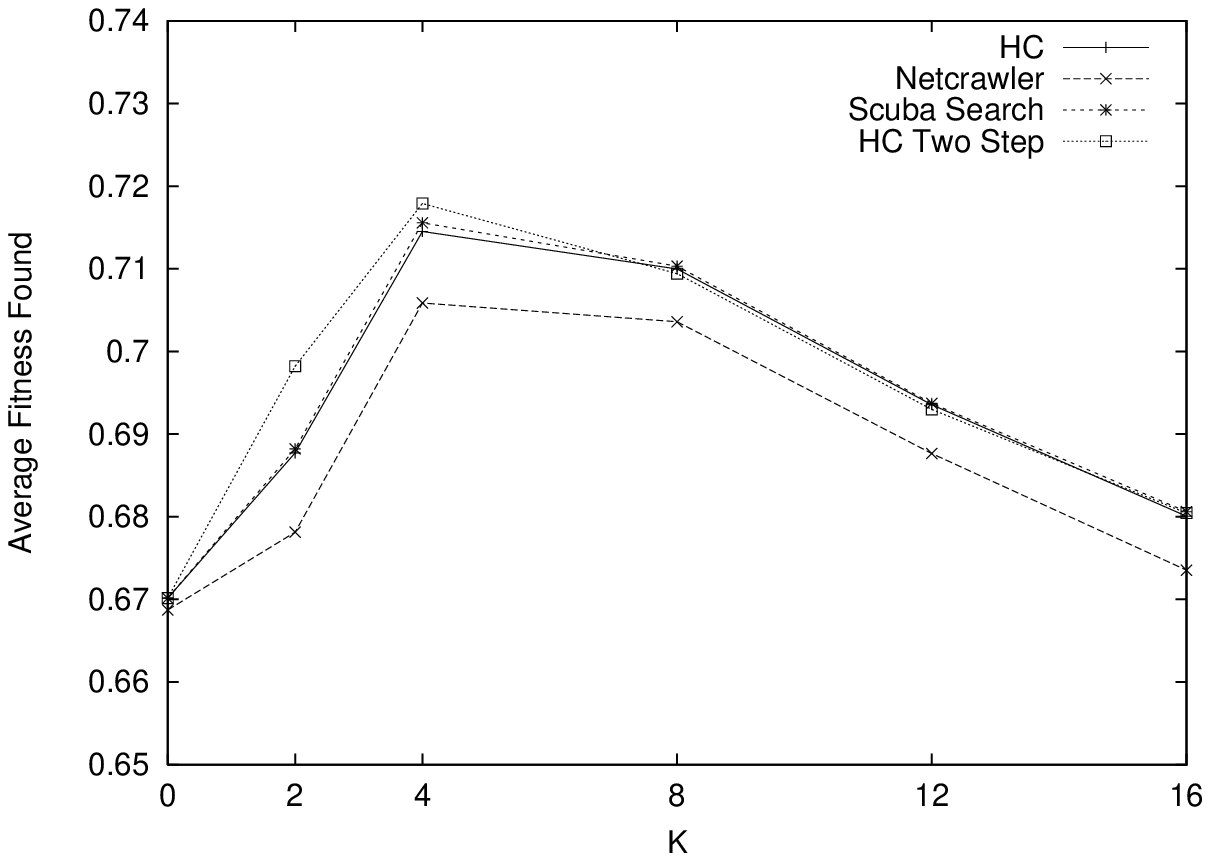,width=230pt,height=180pt} \\
(c) &
(d) \\
\end{tabular}
\end{center}
\caption{Average fitness found on $NKq$-landscapes as function of $K$, for $N=64$ and $q=2$ (a), $q=3$ (b), $q=4$ (c), $q=100$ (d)}
\label{fig_fit_q}
\end{figure*}

Figure \ref{fig_fit_q} shows the average fitness  found respectively by each of the four heuristics as a function of the epistatic parameter $K$ for different values of the neutral parameter $q$. In the presence of neutrality, according to the average fitness, \textit{Scuba Search} outperforms \textit{Hill Climbing}, \textit{Hill Climbing two steps} and \textit{Netcrawler}. Let us note that with high neutrality ($q=2$ and $q=3$), difference is still more significant. Without neutrality ($q=100$) all the heuristics are nearly equivalent, except \textit{Netcrawler}.
\par
The two heuristics adapted to neutral landscape, Scuba Search and Netcrawler, have on average better fitness value for $q=2$ and $q=3$ than hill climbing heuristics. These heuristics benefit in NKq-landscapes from the neutral paths to reach the highest peaks.\\

\subsection{Evaluation cost}

Table \ref{tab_eval} shows the number of evaluations for the different heuristics except for Netcrawler. For this last heuristic the number of evaluations is constant whatever $K$ and $q$ values are; and we get the smallest evaluation cost from all the heuristics ($maxStep=300$ evaluations). For all the other heuristics, the number of evaluations decreases with $K$. The evaluation cost decreases as ruggedness increases. For $HC$ and $HC2$, the evaluation cost increases with $q$. For $HC$ and $HC2$, more neutral the landscape is, smaller the evaluation cost. Conversely, for $SS$ the cost decreases with $q$. 
At each step the number of evaluations is $N$ for $HC$ and $\frac{N(N-1)}{2}$ for $HC2$. So, the cost depends on the length of adaptive walk of $HC$ and $HC2$ only. The evaluation cost of $HC$ and $HC2$ is low when local optima are nearby (i.e. in rugged landscapes).
For $SS$, at each step, the number of evaluations is $(1 + Degn(s))N$ which decreases with neutrality. So, the number of evaluations depends both on the number of steps in $SS$ and on the neutral degree. The evaluation cost of $SS$ is high in neutral landscape.

\begin{table}
\caption{Average number of evaluations on $NKq$-landscape with $N=64$}

\begin{tabular}{|l|c||c|c|c|c|c|c|}
\hline
 & & \multicolumn{6}{|c|}{K} \\ 
\cline{3-8}
& q & 0 & 2 & 4 & 8 & 12 & 16 \\
\hline \cline{3-8}
$HC$      &   & 991 & 961 & 807 & 613 & 491 & 424 \\
$SS$      & 2 & 35769 & 23565 & 15013 & 8394 & 5416 & 3962 \\
$HC2$ &   & 29161 & 35427 & 28038 & 19192 & 15140 & 12374 \\
\hline
$HC$      &   & 1443 & 1159 & 932 & 694 & 546 & 453 \\
$SS$      & 3 & 31689 & 17129 & 10662 & 6099 & 3973 & 2799 \\
$HC2$ &   & 42962 & 37957 & 29943 & 20486 & 15343 & 12797 \\
\hline
$HC$      &   & 1711 & 1317 & 1079 & 761 & 614 & 500 \\
$SS$      & 4 & 22293 & 9342 & 5153 & 2601 & 1581 & 1095 \\
$HC2$ &   & 52416 & 44218 & 34001 & 22381 & 18404 & 14986 \\
\hline
$HC$      &     & 2102 & 1493 & 1178 & 832 & 635 & 517 \\  
$SS$      & 100 & 4175 & 1804 & 1352 & 874 & 653 & 526 \\  
$HC2$ &     & 63558 & 52194 & 37054 & 24327 & 18260 & 15271 \\ 
\hline
\end{tabular}	
\label{tab_eval}
\end{table}

\section{Analysis}

According to the average fitness found, Scuba Search outperforms the others heuristics on the $NKq$ fitness landscapes.
However, it should be wondered whether efficiency of Scuba Search does have with the greatest number of evaluations. 
The neutral search process of Netcrawler and Scuba marked by dissimilarity. We will analyze their own strategy.

\subsection{Exploration neighborhood size}
The number of evaluations for Scuba Search is greater than the one for $HC$ or for Netcrawler. But it lesser than the one for $HC2$. This last heuristic realizes a larger exploration of the neighborhood than $SS$: it pays attention to neighbors with same fitness and all the neighbors of the neighborhood too. However the average fitness found is less good than the one found by $SS$. So, the number of evaluations is not sufficient to explain good performance of $SS$. Whereas there is premature convergence towards local optima with $HC2$, $SS$ realizes a better compromise between exploration and exploitation by examining neutral neighbors.\\


\subsection{Neutral search process}

Difference between the two neutral heuristics can be explained analyzing the respective neutral search process. 
\par
For Netcrawler, as mutation is uniform on the neighborhood, the probability of neutral mutation is proportional to the \textit{neutral degree}: $P_{neutralMut}(s) = \frac{Degn(s)}{N}$. Neutral mutation depends  on the neutral degree only. 
If a mutation is neutral, no way is preferred, all the neutral paths have the same probability to be chosen. The Netcrawler promotes {\it neutral drift}. 
\par
\begin{figure*}[!tb] 
\begin{center}
\begin{tabular}{cc} 
\psfig{figure=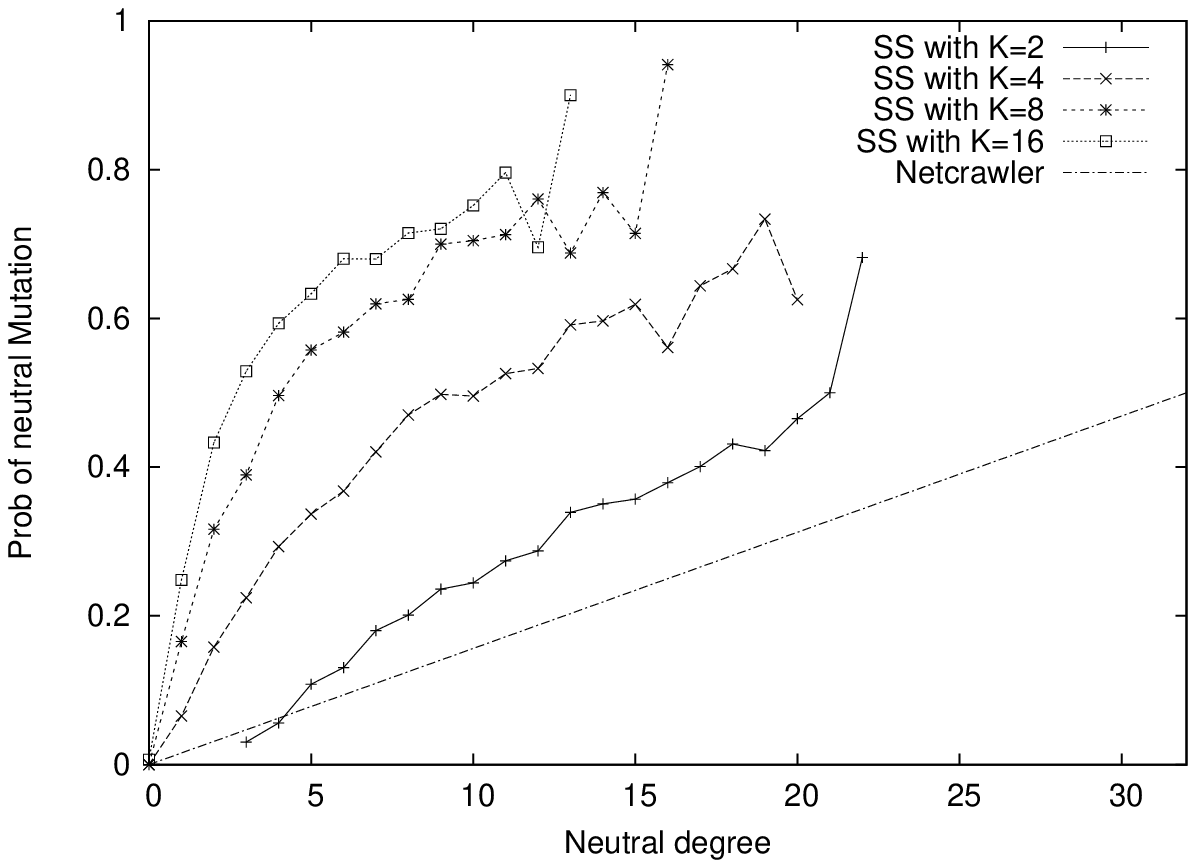,width=230pt,height=180pt}
&
\psfig{figure=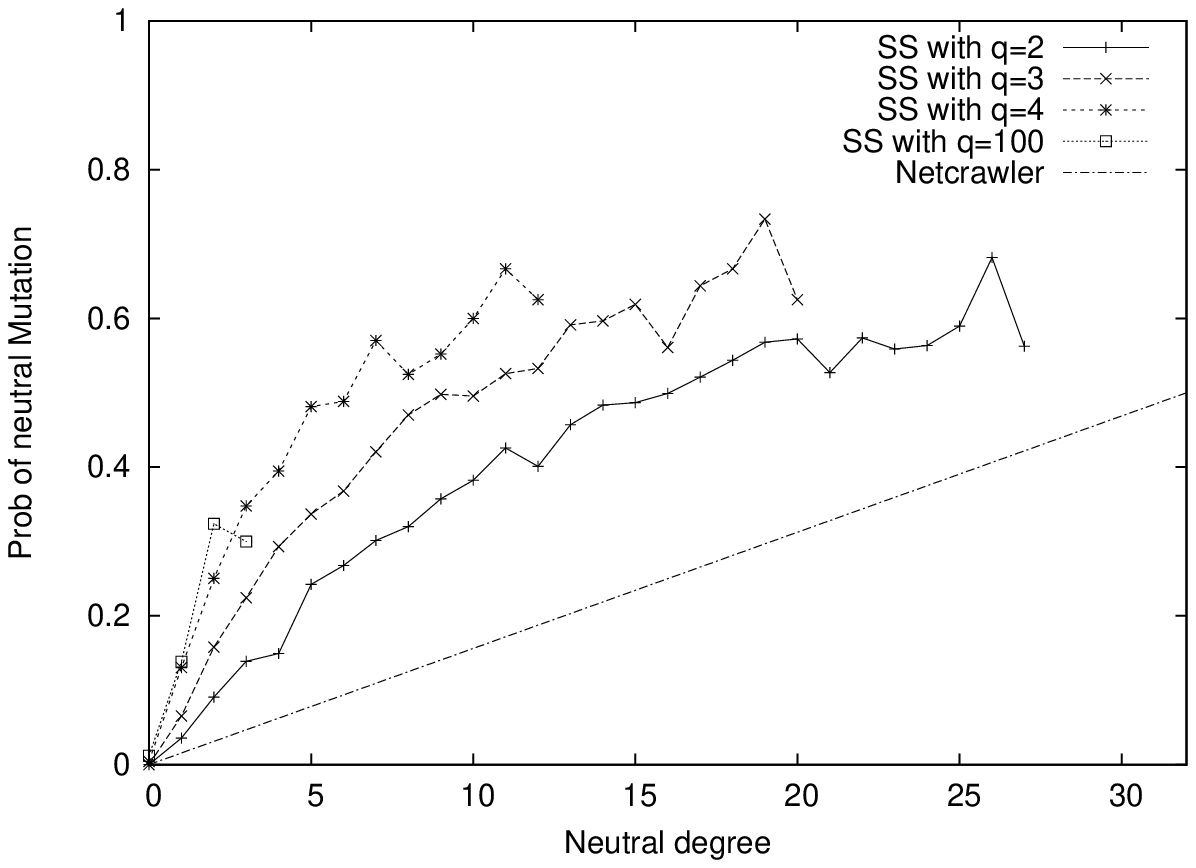,width=230pt,height=180pt} \\

(a) & (b) \\
\end{tabular}
\end{center}
		
	\caption{Probability of neutral mutation for Scuba Search and Netcrawler as a function of neutral degree for $N=64$, $q=3$ (a) and $K=4$ (b)}
	\label{fig_prMutN}

\end{figure*}

For Scuba Search, the probability of neutral mutation increases with the neutral degree (see fig. \ref{fig_prMutN}); without however be proportional. This probability is larger for $SS$ than for Netcrawler and depends on parameters $q$ and $K$ also. Contrary to our intuition about neutral mutation, the probability of neutral move decreases as neutrality increases. The probability of neutral mutation for one given neutral degree increases with $q$ and $K$ as well. 
$SS$ performs a neutral move from genotype $s$ to $s^{'}$ if $s^{'}$ have a strictly greater evolvability than $s$. When neutrality is large, all the neutral neighbors have a strong probability to have the same evolvability. In other words, neutral networks are few connected to neutral sets and then neutral moves are not frequent. When neutrality is a little more important, evolvability of neutral neighbors become quite different and then $SS$ have a larger probability to move while maintaining the same fitness.

\par
\begin{figure}
\centerline{\psfig{figure=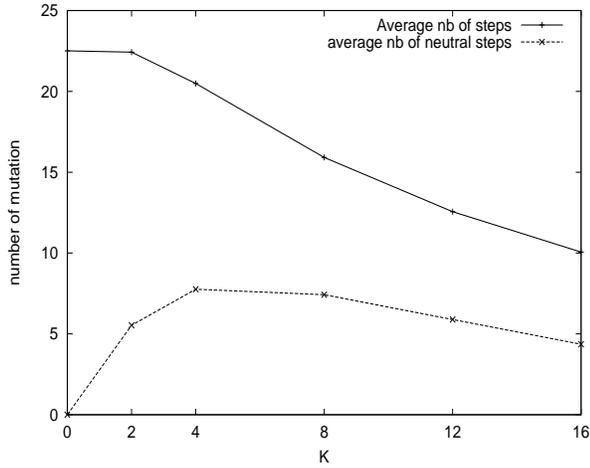,width=230pt,height=180pt}}
			\caption{Number of steps ($gateCount+flatCount$) and neutral mutations ($flatCount$) for Scuba Search Heuristic on $NKq$-landscape with $N=64$ and $q=3$ as a function of $K$.}
			\label{fig_mutMutN_q3}
\end{figure}

Figure \ref{fig_mutMutN_q3} shows the average number of steps ($gateCount + flatCount$) and the average number of neutral mutations ($flatCount$) as a function of the epistatic parameter K computing by the scuba heuristic. The total number of mutations (steps) decreases as epistasis ($K$) increases. The maximum number of neutral mutations is reached for intermediate degree of epistasis ($K=4$); and then the number of neutral moves decreases, but as the probability of neutral move increases, the number of steps decreases faster than the number of neutral moves. So, $SS$ can keep a high number of neutral mutations.

\commentaire{
\begin{figure}
\centerline{\psfig{figure=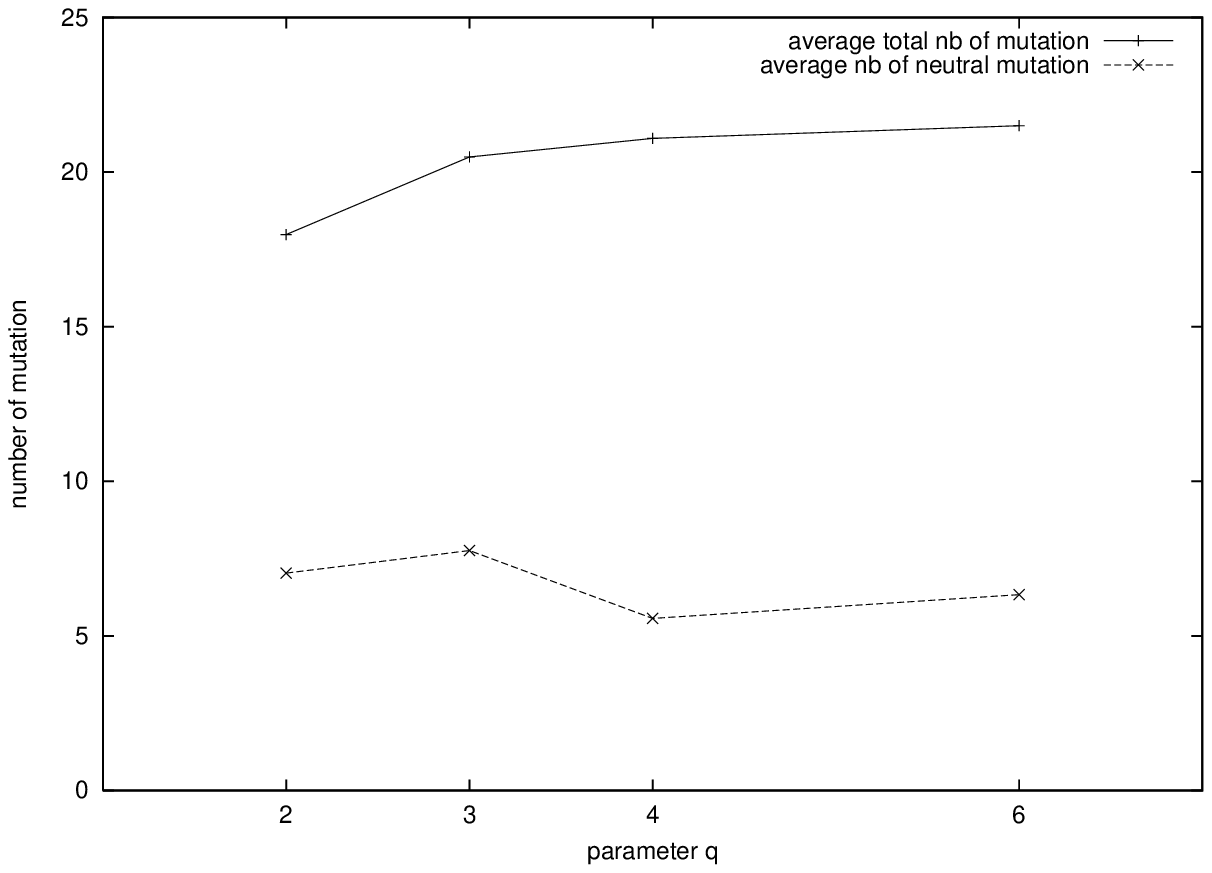,width=160pt,height=140pt}}
			\caption{Number of mutation and neutral mutation for scuba Search Heuristic on $NKq$-landscape with $N=64$ and $K=4$ as a function of $q$}
			\label{fig_mutMutN_K4}
\end{figure}
}

To enlighten differences between the search processes, we map the landscape onto a 2-dimensional space.  This technique inspired by Layzeel~\cite{LAY:Vis} allows to visualize the pathways within a search space that a heurictic is likely to follow. The space is transformed into a connected graph. Vertices are connected if their Hamming distance from each other is one. According to the heuristic used, $HC$, $SS$ or $NC$, salient edges are picked out, and the rest discarded. Each genotype (vertice) is shaded according to its fitness\footnote{In our example, low-fitness genotypes are black}. This allows to "see" all the neutral sets. To illustrate $HC$, $SS$, $NC$, and $HC2$ potential dynamics, a small NKq-landscape, with intermediate epistasis and high neutrality, is considered ($N=5, K=2, q=2$). 

\begin{figure}
\centerline{\psfig{figure=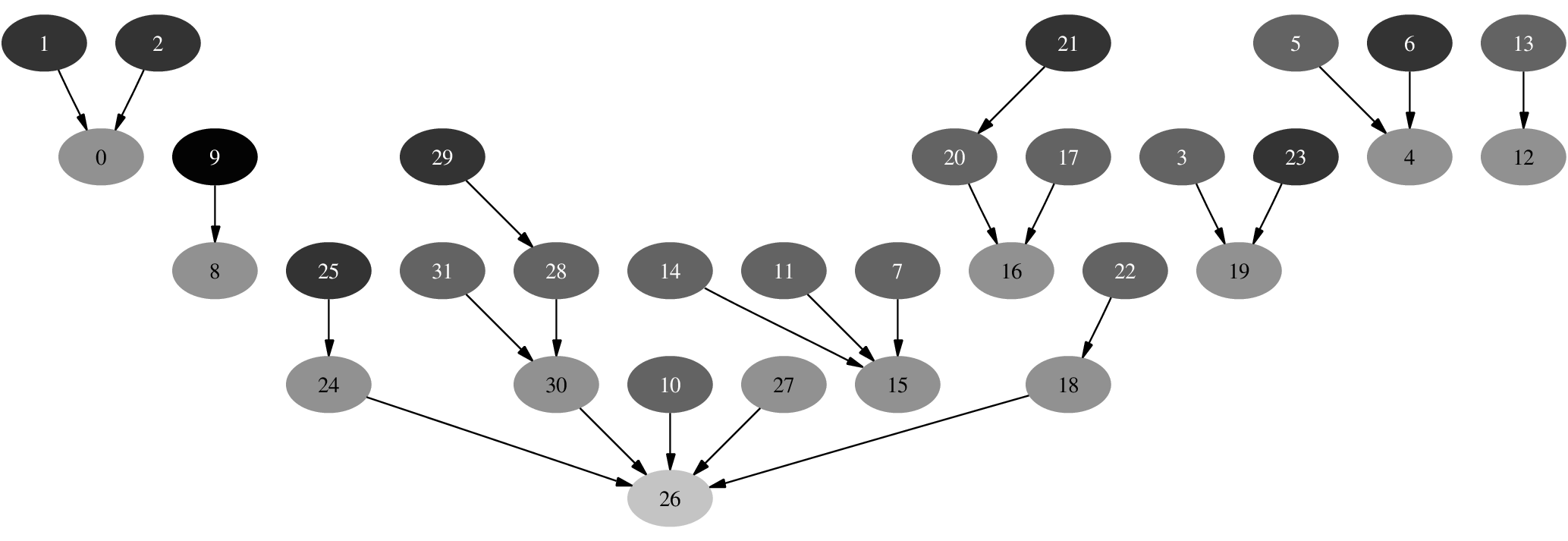,width=250pt,height=140pt}}
			\caption{\textit{Hill Climbing} paths. NKq landscape represented as a connected graph (N=5, K=2, q=2). According to the fitness function $f$, each arrow connects one individual to the fittest genotype in its neighborhood.}
			\label{fig_hc_graph}
\end{figure}

Figure \ref{fig_hc_graph} shows all the \textit{Hill Climbing} paths through the landscape. According to the fitness function $f$, each arrow connects one individual to one of the fittest genotypes in its neighborhood. We can see eight local optima as well the corresponding basins of attraction for this landscape.

\begin{figure}
\centerline{\psfig{figure=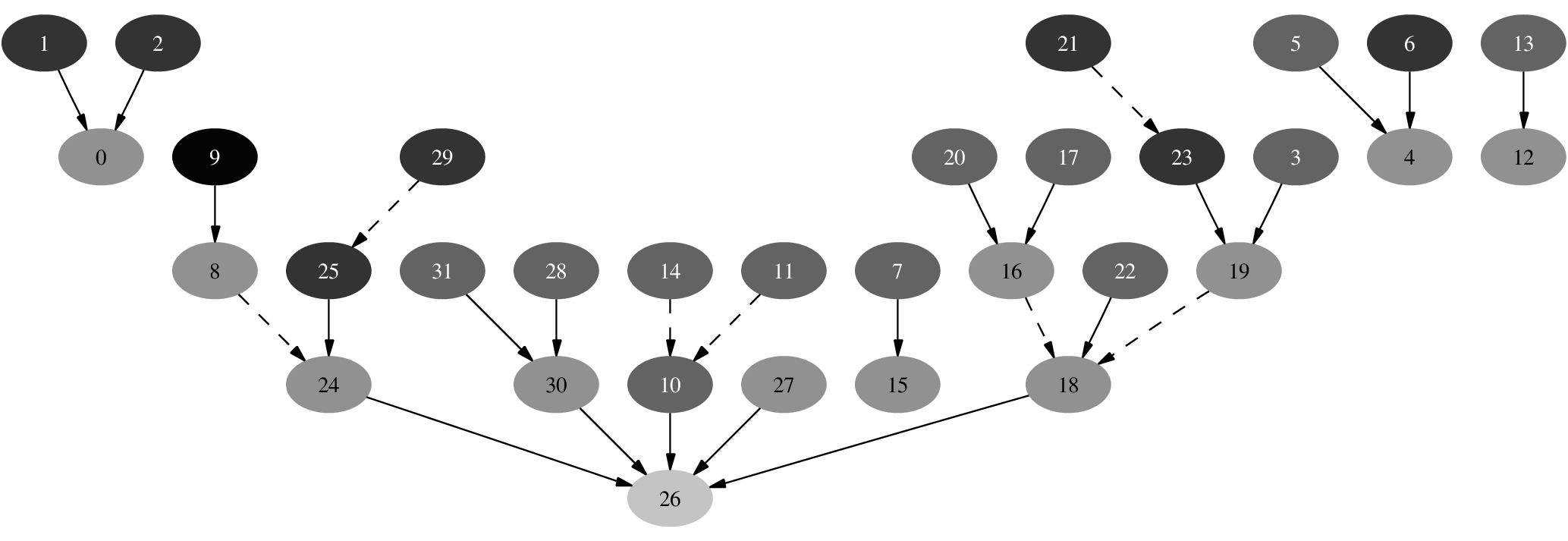,width=250pt,height=140pt}}
			\caption{\textit{Scuba} paths. NKq landscape represented as a connected graph (N=5, K=2, q=2). According to the fitness function $f$, each solid arrow connects one individual to the fittest genotype in its neighborhood. According to the evolvability function $evol$, each dotted arrow connects one individual to the fittest genotype in its neutral neighborhood.}
			\label{fig_scuba_graph}
\end{figure}

Figure \ref{fig_scuba_graph} shows all the \textit{Scuba Search} paths. According to the fitness function $f$, each solid arrow connects one individual to one of the fittest genotypes in its neighborhood. According to the evolvability function $evol$, each dotted arrow connects one individual to one of the fittest genotypes in its neutral neighborhood. The choice between neutral path and $HC$ path is specified by the scuba algorithm (see algorithm \ref{algoScuba}). There are two types of change from $HC$ to $SS$. First, some new neutral paths appear (see for example the path from node 8 to node 24). Second, some $HC$ paths can be replaced by a neutral path (see for instance the new neutral path from node 11 to node 10). Globally the number of local optima and basins of attraction tends to reduce (equal to five here).
\par
\begin{figure}
\centerline{\psfig{figure=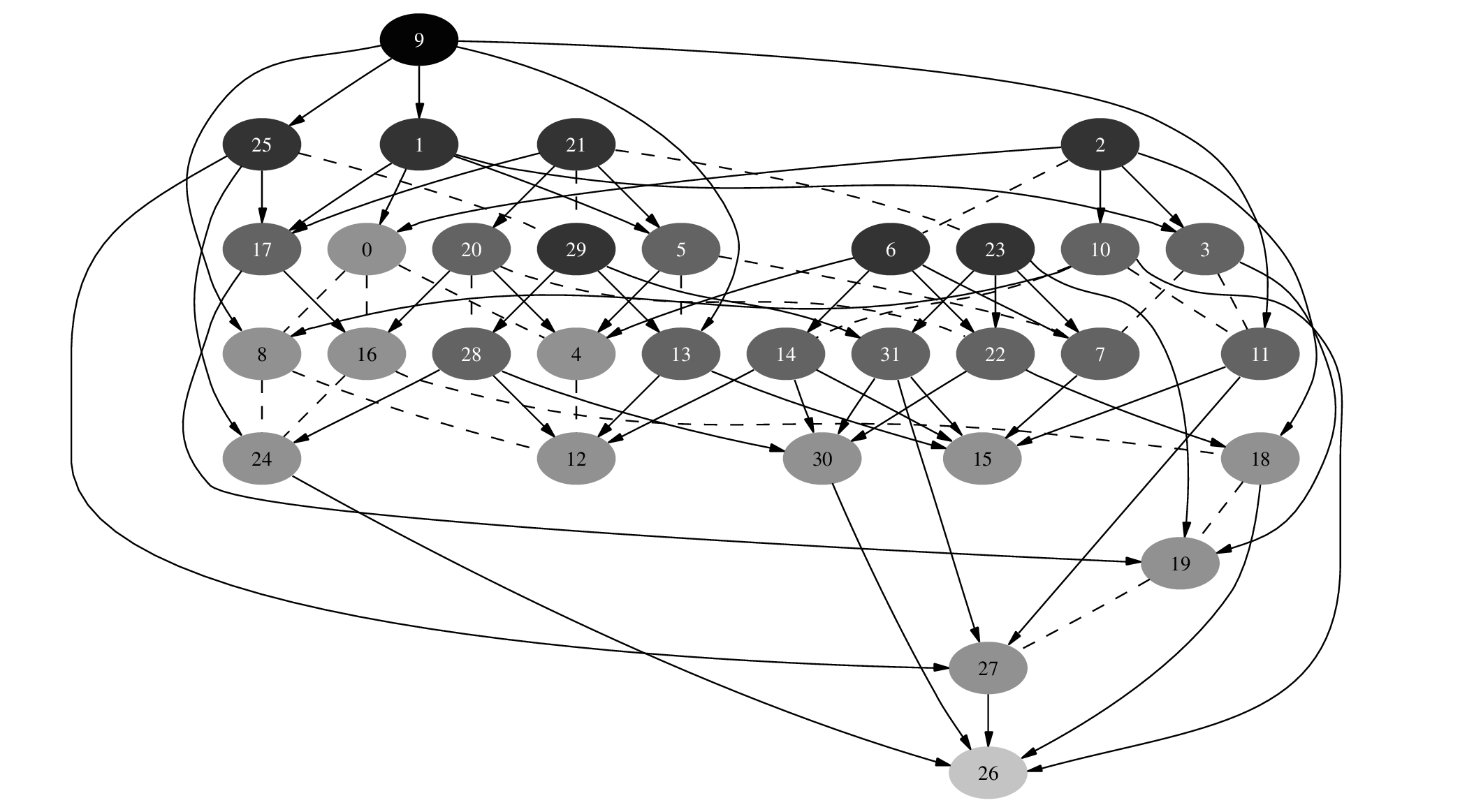,width=280pt,height=180pt}}
			\caption{\textit{Netcrawler} paths. NKq landscape represented as a connected graph (N=5, K=2, q=2). According to the fitness function $f$, each solid arrow connects one individual to a fitter genotype in its neighborhood. Dotted arrows connect neighbors with the same fitness.}
			\label{fig_nc_graph}
\end{figure}

According to the fitness function $f$, each solid arrow in figure \ref{fig_nc_graph} connects one individual to a strict fitter genotype in its neighborhood and dotted arrows connect two neighbors with the same fitness. So, a \textit{Netcrawler Search} path is a subgraph of this graph. 
\par
\begin{figure}
\centerline{\psfig{figure=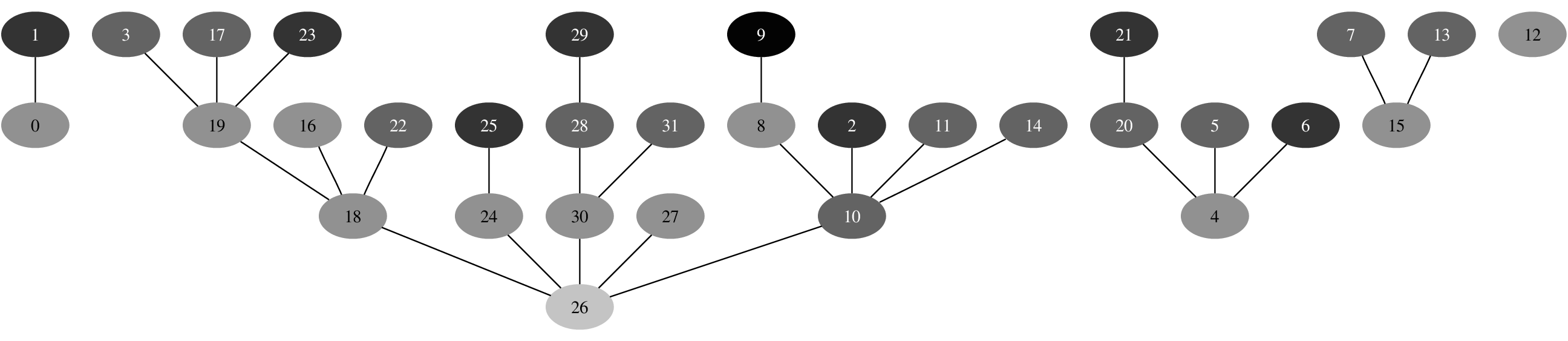,width=250pt,height=100pt}}
			\caption{\textit{Hill Climbing two steps} paths. NKq landscape represented as a connected graph (N=5, K=2, q=2). According to the fitness function $f$, each arrow connects one individual to one of its neighbors which can access to the fittest genotype in its extended neighborhood (size 2).}
			\label{fig_hc2_graph}
\end{figure}

Figure \ref{fig_hc2_graph} shows all the \textit{Hill Climbing two steps} paths. According to the fitness function $f$, each solid arrow connects one individual to genotype as defined in algorithm \ref{algoHCDeuxPas}. As might be expected, compared to $HC$, the number of local optima is smaller (equal to five). But our experiments have shown that coarsely increasing neighborhood size is not sufficient to improve performance. Comparison between $HC2$ and $SS$ performances on NKq-landscape suggests that exploiting neutrality is a better way to guide search heuristics. 

\section{Discussion}

\subsection{Limit of the Scuba search}
The main idea behind Scuba Search heuristic is to try to explicitly optimize evolvability on a neutral network before performing a qualitative step using a local search heuristic. Optimized evolvability needs evolvability not to be constant on a neutral network. For example in the well-known Royal-Road landscape, proposed by Mitchell et al.~\cite{MIT-FOR-HOL:92}, a high degree of neutrality leads evolvability to be constant on each neutral network. A way to reduce this drawback is to modify the neighborhood structure ${\cal V}$ induced by the choice of per-sequence mutation rate; for example we could use the per-sequence mutation rate suggested by Barnett~\cite{barnett01netcrawling}.

\subsection{Generic Scuba Search}
As previously shown, the evaluation cost of $SS$ is relatively large. In oder to reduce this cost, one solution would be to choose a "cheaper" definition for evolvability: for example, the best fitness of $n$ neighbors randomly chosen or the first fitness of neighbor which improves the fitness of the current genotype. Another solution would be to change either the local search heuristic which optimizes evolvability or the one which allows jumping to a fitter solution. For instance, we could use Simulated Annealing~\cite{KIR-GEL-VEC:83} or Tabu search~\cite{GLO:89} to optimize neutral network then jump to the first improvement meets in the neighborhood.

More generally, the Scuba Search heuristic could be extended in three ways:
\begin{itemize}
\item
evolvability definition,
\item
local search heuristic to optimize neutral network,
\item
local search heuristic to jump toward fitter solution.
\end{itemize}
From these remarks we propose to defined the {\it Generic Scuba Search} (alg. \ref{GenAlgoScuba}).

\begin{algorithm}
\caption{Generic Scuba Search}
\label{GenAlgoScuba}

\begin{algorithmic}
\STATE flatCount $\leftarrow$ 0, gateCount $\leftarrow$ 0
\STATE Choose initial solution $s \in \cal S$
\REPEAT
	\WHILE{terminal condition$_1$ not met}
		\STATE $s \leftarrow$ Improve$_1(s, evol, {\cal V}n(s))$
		\STATE flatCount $\leftarrow$ flatCount + 1
	\ENDWHILE
	\STATE $s \leftarrow$ Improve$_2(s, f, {\cal V}(s))$
	\STATE gateCount $\leftarrow$ gateCount+1
\UNTIL{terminal condition$_2$ met}
\end{algorithmic}
\end{algorithm}

\section{Conclusion and Perspectives}

This paper represents a first step demonstrating the potential interest in using scuba search heuristic. According to the average fitness found, $SS$ outperforms hill climbing heuristics and netcrawler on the $NKq$ fitness landscapes. Comparison with $HC2$ algorithm has shown that $SS$ efficiency does not have with the number of evaluations only. Mapping the landscape onto 2-dimensional space allows qualitative difference between neutral search processes to emerge.
\par
When neutrality is too high the scuba search stops in the 'middle' of neutral networks. 
In future work, we would like to replace Improve$_1$ heuristic by a tabu search allowing to drift on neutral networks.
\par
As our implementation uses hill climbing as Improve$_1$ and Improve$_2$ heuristics, $SS$ is more to be compare to a local search heuristic without deleterious mutation. Thus using Tabu Search on neutral landscape it could be useful to replace $HC$ by $SS$. At each iteration of tabu search we choose a new better solution which is not in the tabu list. This better solution is a neutral neighbor if its evolvabilty is greater than evolvabilty of current solution. If not, it is the fittest neighbor. 
\par
Performance of heuristics adapted to neutral landscape depends on 
the difficulty to optimize neutral networks and the connectivity between networks. It would be interesting to have a measure 
of neutral networks optimization difficulty. Moreover if we are able to measure the difficulty to jump from one neutral network to another one then we can compare the efficiency of neutral exploration.
\par
Last but not least we obviously have to study scuba search on other problems than $NKq$-landscapes, in particular on real-world fitness landscapes where neutrality already naturaly exists.

\bibliography{Biblio}
\bibliographystyle{unsrt} 

\end{document}